%% file: main.tex
\documentclass[10pt,twocolumn,letterpaper]{article}

\usepackage{cvpr}
\usepackage{times}
\usepackage{epsfig}
\usepackage{graphicx}
\usepackage{amsmath}
\usepackage{amssymb}
\usepackage{textcomp}

\usepackage{xspace}
\usepackage{color}
\usepackage{booktabs}
\usepackage{subcaption}
\usepackage{bbm}
\usepackage{microtype}
\usepackage{tikz}
\usetikzlibrary{positioning}
\usetikzlibrary{arrows.meta}
\usetikzlibrary{calc}
\tikzset{}
\usepackage{pgfplots}
\pgfplotsset{width=2.5cm,compat=newest}

\usepackage[T1]{fontenc}
\usepackage[utf8]{inputenc}

\usepackage[breaklinks=true,bookmarks=false]{hyperref}

\input{custom_cmds}

\input{notation}

\makeatletter
\renewcommand{\paragraph}{%
  \@startsection{paragraph}{4}%
  {\z@}{0.5ex \@plus 1ex \@minus .2ex}{-1em}%
  {\normalfont\normalsize\bfseries}%
}
\makeatother

\usepackage{enumitem}
\setitemize[0]{leftmargin=10pt,itemsep=0pt,topsep=2pt,parsep=0pt}
\setenumerate[0]{leftmargin=16pt,itemsep=0pt,topsep=2pt,parsep=0pt}

\cvprfinalcopy 

\def\cvprPaperID{4914} 

\ifcvprfinal\fi
\begin{document}

\title{A Parametric Top-View Representation of Complex Road Scenes}

\author{Ziyan Wang$^{1}$\thanks{Work done during an internship at NEC Laboratories America.} $\quad$ Buyu Liu$^{2}$ $\quad$ Samuel Schulter$^{2}$ $\quad$ Manmohan Chandraker$^{2,3}$ \\
$^1$Carnegie Mellon University $\quad$ $^2$NEC Laboratories America $\quad$ $^3$UC San Diego
}
%
%

\maketitle
\thispagestyle{empty}

\input{sections/abstract}
\input{sections/intro}
\input{sections/related_work}
\input{sections/method}
\input{sections/experiments}
\input{sections/conclusion}

{\small
\bibliographystyle{ieee}
\bibliography{myshortstrings,egbib}
}

\end{document}

%% file: custom_cmds.tex
\newlength\tikzfigwidth
\newlength\tikzfigheight

%% file: notation.tex
\def\txtBin{\textrm{b}}
\def\txtMc{\textrm{m}}
\def\txtReg{\textrm{c}}

\def\sa{\Theta}
\def\saBin{\sa_{\txtBin}}
\def\saMc{\sa_{\txtMc}}
\def\saReg{\sa_{\txtReg}}
\def\saBinIdx{\sa_{\txtBin{},i}}
\def\saMcIdx{\sa_{\txtMc,i}}
\def\saRegIdx{\sa_{\txtReg,i}}
\def\saBinNum{M^{\txtBin}}
\def\saMcNum{M^{\txtMc}}
\def\saRegNum{M^{\txtReg}}
\def\numSaRegBins{K}
\def\saSim{\sa^{\textrm{s}}}
\def\saReal{\sa^{\textrm{r}}}

\def\saBinRealSimIdx{\saBinIdx^{\{\textrm{r,s}\}}}
\def\saMcRealSimIdx{\saMcIdx^{\{\textrm{r,s}\}}}
\def\saRegRealSimIdx{\saRegIdx^{\{\textrm{r,s}\}}}

\def\dsReal{D^{\textrm{r}}}
\def\dsSim{D^{\textrm{s}}}
\def\dsRealSize{N^{\textrm{r}}}
\def\dsSimSize{N^{\textrm{s}}}
\def\dsRealSimSize{N^{\{\textrm{r,s}\}}}

\def\bevmap{\mat{x}}
\def\bevmapH{H}
\def\bevmapW{W}
\def\bevmapC{C}
\def\bevmapFeat{f_{\bevmap}}
\def\bevmapFeatDim{D}
\def\bevmapReal{\bevmap^{\textrm{r}}}
\def\bevmapSim{\bevmap^{\textrm{s}}}
\def\bevmapRealSim{\bevmap^{\textrm{r,s}}}
\def\bevmapFeatReal{f_{\bevmapReal}}
\def\bevmapFeatSim{f_{\bevmapSim}}
\def\bevmapFeatRealSim{f_{\bevmapRealSim}}

\def\nnFull{f}
\def\nnFeat{g}
\def\nnFeatReal{g_{\textrm{r}}}
\def\nnFeatSim{g_{\textrm{s}}}
\def\nnAttr{h}

\def\nnFeatWgts{\gamma_{\nnFeat}}
\def\nnFeatWgtsReal{\gamma_{\nnFeat}^{\textrm{r}}}
\def\nnFeatWgtsSim{\gamma_{\nnFeat}^{\textrm{s}}}
\def\nnAttrWgts{\gamma_{\nnAttr}}
\def\nnDiscr{d}
\def\nnDiscrWgts{\gamma_{\nnDiscr}}

\def\predSa{\eta}
\def\predSaBin{\predSa_{\txtBin}}
\def\predSaMc{\predSa_{\txtMc}}
\def\predSaReg{\predSa_{\txtReg}}
\def\predSaBinIdx{\predSa_{\txtBin,i}}
\def\predSaMcIdx{\predSa_{\txtMc,i}}
\def\predSaRegIdx{\predSa_{\txtReg,i}}

\def\predSaBinRealSimIdx{\predSaBinIdx^{\{\textrm{r,s}\}}}
\def\predSaMcRealSimIdx{\predSaMcIdx^{\{\textrm{r,s}\}}}
\def\predSaRegRealSimIdx{\predSaRegIdx^{\{\textrm{r,s}\}}}

\def\lossSym{\mathcal{L}} 

\def\lossSup{\lossSym_{\textrm{sup}}}
\def\lossSupReal{\lossSup^{\textrm{r}}}
\def\lossSupSim{\lossSup^{\textrm{s}}}
\def\lossSupRealSim{\lossSup^{\{\textrm{r,s}\}}}
\def\lossAdv{\lossSym_{\textrm{adv}}}

\def\lossWgtReal{\lambda^{\textrm{r}}}
\def\lossWgtSim{\lambda^{\textrm{s}}}
\def\lossWgtAdv{\lambda^{\textrm{adv}}}

\def\lossBCE{\textrm{BCE}}
\def\lossCE{\textrm{CE}}
\def\lossEllOne{\ell_1}

\def\realspace{\mathbb{R}} 

\newcommand{\mat}[1]{\ensuremath{\mathbf{#1}}}
 
\newcommand{\chis}[2][non]{\ensuremath{\chi^2 \ifthenelse{\equal{#1}{non}}{}{ \left(#1,#2\right)}}} 
\newcommand{\Gammaf}[1][non]{\ensuremath{\Gamma\ifthenelse{\equal{#1}{non}}{}{ \left( #1 \right)}}} 

\newcommand{\degree}[1][non]{\ensuremath{\ifthenelse{\equal{#1}{non}}{^\circ}{#1^\circ}}} 
\LetLtxMacro{\oldsqrt}{\sqrt}
\renewcommand{\sqrt}[1][\ ]{%
  \def\DHLindex{#1}\mathpalette\DHLhksqrt}
\def\DHLhksqrt#1#2{%
  \setbox0=\hbox{$#1\oldsqrt[\DHLindex]{#2\,}$}\dimen0=\ht0
  \advance\dimen0-0.2\ht0
  \setbox2=\hbox{\vrule height\ht0 depth -\dimen0}%
  {\box0\lower0.71pt\box2}}



%% file: sections/abstract.tex
\begin{abstract}
In this paper, we address the problem of inferring the layout of complex road scenes given a single camera as input.  To achieve that, we first propose a novel parameterized  model of road layouts in a top-view representation, which is not only intuitive for human visualization but also provides an interpretable interface for higher-level decision making.  Moreover, the design of our top-view scene model allows for efficient sampling and thus generation of large-scale simulated data, which we leverage to train a deep neural network to infer our scene model's parameters.  Specifically, our proposed training procedure uses supervised domain-adaptation techniques to incorporate both simulated as well as manually annotated data.  Finally, we design a Conditional Random Field (CRF) that enforces coherent predictions for a single frame and encourages temporal smoothness among video frames.  Experiments on two public data sets show that: (1) Our parametric top-view model is representative enough to describe complex road scenes; (2) The proposed method outperforms baselines trained on manually-annotated or simulated data only, thus getting the best of both; (3) Our CRF is able to generate temporally smoothed while semantically meaningful results.

\end{abstract}


%% file: sections/intro.tex
\section{Introduction}
\label{sec:intro}
Understanding complex layouts of the 3D world is a crucial ability for applications like robot navigation, driver assistance systems or autonomous driving.
Recent success in deep learning-based perception systems enables pixel-accurate semantic segmentation~\cite{sam:RotaBulo18a,sam:Chen18a,sam:Zhao17a} and (monocular) depth estimation~\cite{sam:Godard17a,sam:Laina16a,sam:Xu18a} in the perspective view of the scene.
Other works like~\cite{sam:Guo12a,sam:Schulter18a,sam:Sengupta12a} go further and reason about occlusions and build better representations for 3D scene understanding.
The representation in these works, however, is typically non-parametric, \ie, it provides a semantic label for a 2D/3D point of the scene, which makes higher-level reasoning hard for downstream applications.

\begin{figure}\centering
  \includegraphics[width=1.0\columnwidth]{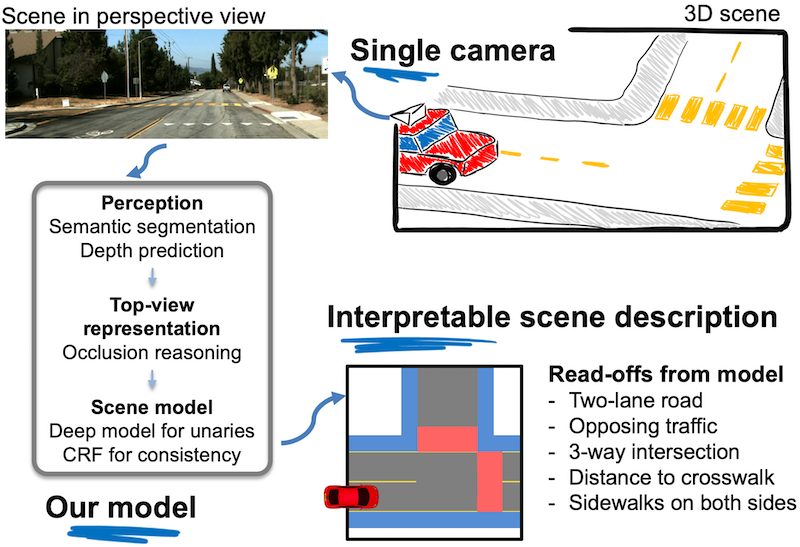}
  \vspace{-0.65cm}
  \caption{
  Our goal is to infer the layout of complex driving scenes from a single camera.  Given a perspective image (top left) that captures a 3D scene, we predict a rich and interpretable scene description (bottom right), which represents the scene in an occlusion-reasoned semantic top-view.
  }
  \label{fig:teaser}
\end{figure}

In this work, we focus on understanding driving scenarios and propose a rich \emph{parameterized} model describing complex road layouts in a top-view representation (Fig.~\ref{fig:teaser} and Sec.~\ref{sec:scene_model}).  The parameters of our model describe important scene attributes like the number and width of lanes, and the existence of, and distance to various types of intersections, crosswalks and sidewalks.  An explicit model of such parameters is beneficial for higher-level modeling and decision making as it provides a tangible interface to the real world.  In contrast to prior art~\cite{sam:Geiger14a,sam:Kunze18a,sam:Mattyus16a,sam:Schulter18a,sam:Seff16a,sam:Sengupta12a}, our proposed scene model is richer, fully parameterized and can be inferred from a single camera input with a combination of deep neural networks and a graphical model.

However, training deep neural networks requires large amounts of training data.  Although annotating the scene attributes of our model for real RGB images is possible, it is also costly to do at a large-scale and, more importantly, extremely difficult for certain scene attributes. While the existence of a crosswalk is a binary attribute and is easy to annotate, annotating the exact width of a side road requires the knowledge of scene geometry, which is hard when only given a perspective RGB image.  We thus propose to leverage simulated data.  However, in contrast to rendering photo-realistic RGB images, which is a difficult and time-consuming task~\cite{sam:Richter17a,sam:Richter16a}, we propose a scene model that allows for efficient sampling and render semantic top-view representations that obviate expensive illumination modeling or occlusion reasoning. 

Given simulated data with accurate and complete annotations, as well as real images with potentially noisy and incomplete annotations, we propose a hybrid training procedure leveraging both sources of information.
Specifically, our neural network design involves (i) a feature extractor that aims to leverage information from both domains, simulated and real semantic top-views from \cite{sam:Schulter18a} (see Fig.~\ref{fig:top_view_examples_real_and_sim}), and (ii) a domain-agnostic classifier of scene parameters.
At test time, we convert a perspective RGB image into a semantic top-view representation using \cite{sam:Schulter18a} and predict our scene model's parameters. Given the individual scene parameter predictions, we further design a graphical model (Sec.~\ref{sec:graphical_model}) that captures dependencies among scene attributes in single images and enforces temporal consistency across a sequence of frames.
We validate our idea on two public driving data sets, KITTI~\cite{sam:Geiger13a} and NuScenes~\cite{sam:NuScenes18a} (Sec.~\ref{sec:exps}).  The results demonstrate the effectiveness of the top-view representation, the hybrid training procedure with real and simulated data, and the importance of the graphical model for coherent and consistent outputs.
To summarize, our key contributions are:
\begin{itemize}
\item A \textbf{novel parametric and interpretable model} of complex driving scenes in a top-view representation.
\item A neural network that (i) predicts the parameters from a single camera and (ii) is designed to enable a \textbf{hybrid training approach from both real and synthetic data}.
\item A graphical model that ensures \textbf{coherent and temporally consistent} scene description outputs.
\item \textbf{New annotations} of our scene attributes for the KITTI~\cite{sam:Geiger13a} and NuScenes~\cite{sam:NuScenes18a} data sets\footnote{\url{http://www.nec-labs.com/~mas/BEV}}.
\end{itemize}


%% file: sections/related_work.tex
\section{Related Work}
\label{sec:related}
3D scene understanding is an important task in computer vision with many applications for robot navigation~\cite{sam:Gupta17a},  self-driving~\cite{sam:Geiger14a,sam:Kunze18a}, augmented reality~\cite{sam:Armeni16a} or real estate~\cite{sam:Liu15a,sam:Song18a}.

\paragraph{Scene understanding:} Explicit modeling of the scene is frequently done for indoor applications where strong priors about the layout of rooms can be leveraged~\cite{sam:Armeni16a,sam:Liu15a,sam:Song17a}.
Non-parametric approaches are more common for outdoor scenarios because the layout is typically more complex and harder to capture in a coherent model, with occlusion reasoning often being a primary focus. Due to the natural ability to reflect orders, layered representations~\cite{sam:Guo12a,Tighe_2014_CVPR,sam:Tulsiani18a} have been utilized in scene understanding to reason about geometry and semantics in occluded areas. However, such intermediate representations are not desired for applications where distance information is required. A top-view representation~\cite{sam:Schulter18a,sam:Sengupta12a}, in contrast, is a more detailed description for 3D scene understanding. Our work follows such a top-view representation and aims to infer a parametric model of complex outdoor driving scenes from a single input image.

A few parametric models have been proposed for outdoor environments too. Seff and Xiao~\cite{sam:Seff16a} present a neural network that directly predicts scene attributes from a single RGB image. Although those attributes are automatically acquired from OpenStreetMaps~\cite{OpenStreetMap}, they are not rich enough to fully describe complex road scenes, e.g. curved road with side-roads. A richer model that is capable of handling complex intersections with traffic participants is proposed by Geiger~\etal~\cite{sam:Geiger14a}. To this end, they propose to utilize multiple modalities such as vehicle tracklets, vanishing points and scene flow. Different from their work,
we focus more on scene layouts and propose in Sec.~\ref{sec:scene_model} a richer model in that aspect, including multiple lanes, crosswalks and sidewalks. Moreover, our base framework is able to infer model parameters with a single perspective image as input. A more recent work~\cite{sam:Kunze18a} proposes to infer a graph representation of the road, including lanes and lane markings, from partial segmentations of an image. Unlike our method that aims to handle complex road scenarios, \cite{sam:Kunze18a} focuses only on straight roads.
M\'{a}ttyus~\etal~\cite{sam:Mattyus16a} propose an interesting parametric model of roads with the goal of augmenting existing map data with richer semantics. Again, this model only handles straight roads and requires input from both perspective and aerial images.
Perhaps~\cite{sam:Schulter18a} is the closest work to ours. In contrast to it, we propose a fully-parametric model that is capable of reconstructing complex road layouts.

\paragraph{Learning from simulated data:} Besides the scene model itself, one key contribution of our work is the training procedure that leverages simulated data, where we also utilize tools from domain adaptation~\cite{sam:Ganin16a,sam:Tsai18a}.
While most recent advances in this area focus on bridging domain gaps between synthetic and real RGB images~\cite{sam:Richter17a,sam:Richter16a}, we benefit from the semantic top-view representation within which our model is defined. 
This representation allows efficient modeling and sampling of a variety of road layouts, while avoiding the difficulty of photo-realistic renderings, to significantly reduce the domain gap between simulated and real data.


%% file: sections/method.tex
\begin{figure*}\centering
  \includegraphics[width=1.0\textwidth]{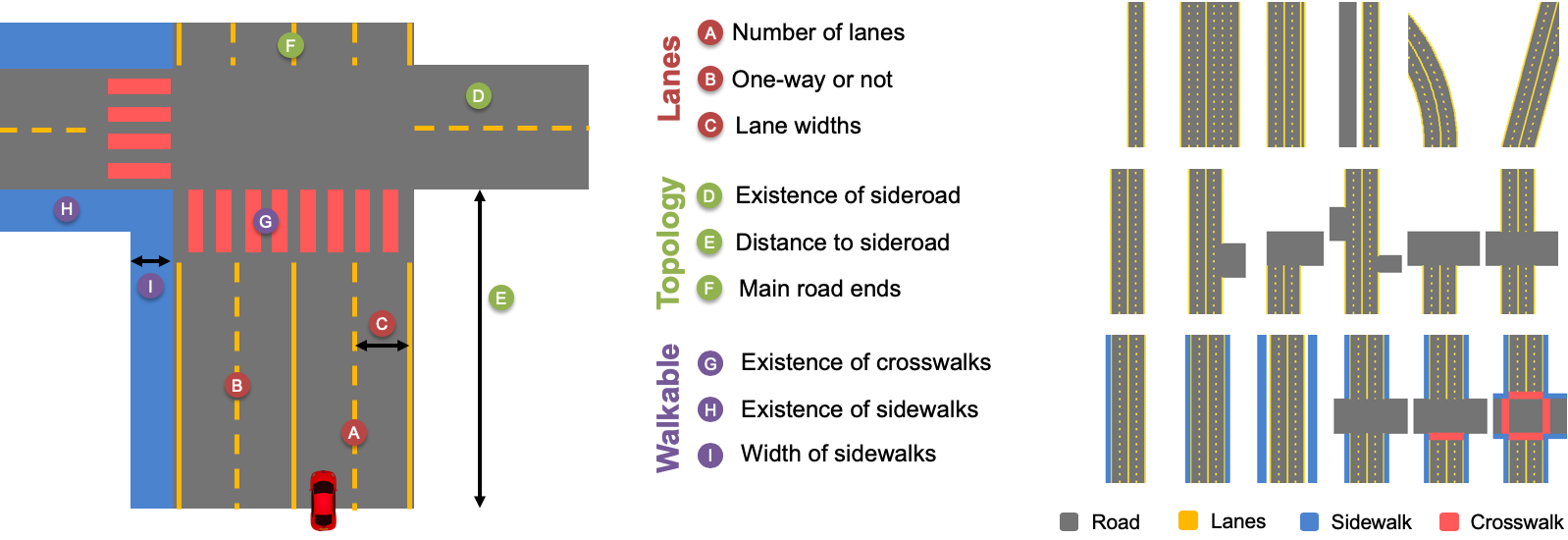}
  \vspace{-0.65cm}
  \caption{Our scene model consists of several parameters that capture a variety of complex driving scenes.  (Left) We illustrate the model and highlight important parameters (A-I), which are grouped into three categories (middle): \emph{Lanes}, to describe the layout of a single road; \emph{Topology}, to model various road topologies; \emph{Walkable}, describing scene elements for pedestrians.  Our model is defined as a directed acyclic graph enabling efficient sampling and is represented in the top-view, making rendering easy.  These properties turn our model into a simulator of semantic top-views.  (Right) We show rendered examples for each of the above groups.  A complete list of scene parameters and the corresponding graphical model is given in the supplementary.
  }
  \label{fig:overview_scene_model}
\end{figure*}

\section{Our Framework}
\label{sec:method}
The goal of this work is to extract interpretable attributes of the layout of complex road scenes from a single camera.  Sec.~\ref{sec:scene_model} presents our first contribution, a parameterized and rich model of road scenes describing attributes like the topology of the road, the number of lanes or distances to scene elements.  The design of our scene model allows efficient sampling and, consequently, enables the generation of large-scale simulated data with accurate and complete annotations.  At the same time, manual annotation of such scene attributes for real images is costly and, more importantly, even infeasible for some attributes, see Sec.~\ref{sec:traindata_real_sim}.  The second contribution of our work, described in Sec.~\ref{sec:train_infer_scene_model}, is a deep learning framework that leverages training data from both domains, real and simulation, to infer the parameters of our proposed scene model.  Finally, our third contribution is a conditional random field (CRF) that enforces coherence between related parameters of our scene model and encourages temporal smoothness for video inputs, see Sec.~\ref{sec:graphical_model}.

\subsection{Scene Model}
\label{sec:scene_model}
Our model describes road scenes in a semantic top-view representation and we assume the camera to be at the bottom center in every frame.  This allows us to position all elements relative to the camera.  On a higher level, we differentiate between the ``main road'', which is where the camera is, and eventual ``side roads''.  All roads consist of at least one lane and intersections are a composition of multiple roads.  Fig.~\ref{fig:overview_scene_model} gives an overview of our proposed model.

Defining two side roads (one on the left and one on the right of the main road) along with distances to each one of them gives us the flexibility to model both 3-way and 4-way intersections.  An additional attribute determines if the main road ends after the intersection, which yields T-intersections.

The main road is defined by a set of lanes, one- or two-way traffic, delimiters and sidewalks.  We also define up to six lanes on the left and right side of the camera, which occupies the ego-lane.  We allow different lane widths to model special lanes like turn- or bike-lanes.  Next to the outer most lanes, optional delimiters of a certain width separate the road from the optional sidewalk.  At intersections, we also model the existence of crosswalks at all four potential sides.  For side roads, we only model their width.
Our final set of parameters $\sa$ is grouped into different types and we count $\saBinNum = 14$ binary variables $\saBin$, $\saMcNum = 2$ multi-class variables $\saMc$ and $\saRegNum = 22$ continuous variables $\saReg$.  The supplemental material contains a complete list of our model parameters. Note that the ability to work with a simple simulator means we can easily extend our scene model with further parameters and relationships.

\begin{figure*}\centering
  \includegraphics[width=1.0\textwidth]{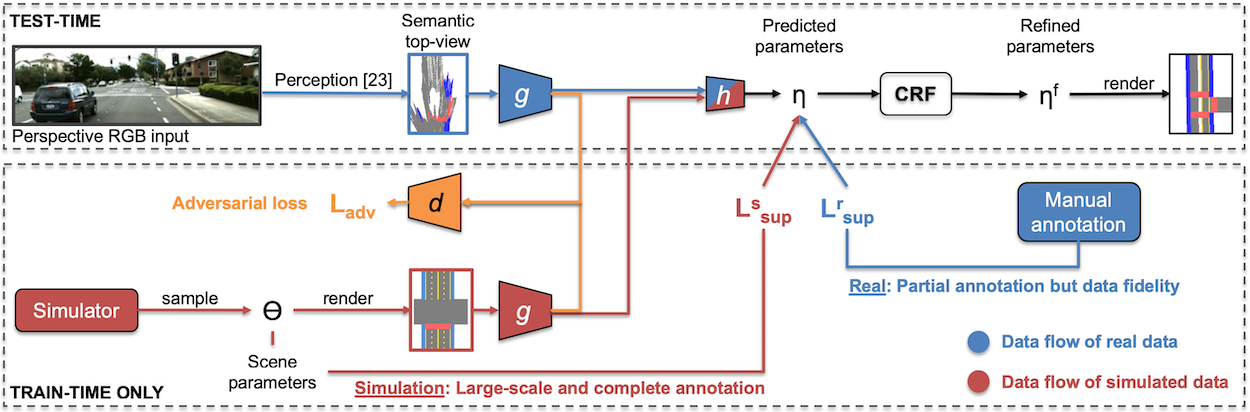}
  \vspace{-0.65cm}
  \caption{\textbf{Overview of our proposed framework:} At train-time, our framework makes use of both manual annotation for real data (blue) and automated annotation for simulated data (red), see Sec.~\ref{sec:traindata_real_sim}. The feature extractors $\nnFeat$ convert semantic top views from either domain into a common representation which is input to $\nnAttr$. An adversarial loss (orange) encourages a domain-agnostic output of $\nnFeat$. At test-time, an RGB image in the perspective view is first transformed into a semantic top-view~\cite{sam:Schulter18a}, which is then used by our proposed neural network (see Sec.~\ref{sec:train_infer_scene_model}), $\nnAttr \circ \nnFeat$, to infer our scene model (see Sec.~\ref{sec:scene_model}). The graphical model defined in Sec.~\ref{sec:graphical_model} ensures a coherent final output.}
  \label{fig:framework_overview}
\end{figure*}

\subsection{Supervision from Real and Simulated Data}
\label{sec:traindata_real_sim}
Inferring our model's parameters from an RGB image requires abundant training data.  Seff and Xiao~\cite{sam:Seff16a} leverage OpenStreetMaps~\cite{OpenStreetMap} to gather ground truth for an RGB image.  While this can be done automatically given the GPS coordinates, the set of attributes retrievable is limited and can be noisy.  Instead, we leverage a combination of manual annotation and simulation for training.

\paragraph{Real data:}
Annotating real images with attributes corresponding to our defined parameters can be done efficiently only when suitable tools are used.  This is particularly true for sequential data because many attributes stay constant over a long period of time.  The supplemental material contains details on our annotation tool and process.  We have collected a data set $\dsReal = \{\bevmapReal, \saReal\}_{i=1}^{\dsRealSize}$ of $\dsRealSize$ samples of semantic top-views $\bevmapReal$ and corresponding scene attributes $\saReal$.  The semantic top-views $\bevmapReal \in \realspace^{\bevmapH \times \bevmapW \times \bevmapC}$, with spatial dimensions $\bevmapH \times \bevmapW$, contain $\bevmapC$ semantic categories ("road", "sidewalk", "lane boundaries" and "crosswalks") and are computed by applying the framework of~\cite{sam:Schulter18a}.  However, several problems arise with real data.  First, ground truth depth is required at a reasonable density for each RGB image to ask humans to reliably estimate distances to scene elements like intersections or crosswalks.  Second, there is always a limit on how much diverse data can be annotated cost-efficiently.  Third, and most importantly, not all desired scene attributes are easy or even possible to annotate at a large-scale, even if depth information is available.  For these reasons, we explore simulation as another source of supervision.

\paragraph{Simulated data:}
Our proposed scene model defined in Sec.~\ref{sec:scene_model} can act as a simulator to generate training data with complete and accurate annotation.  First, by treating each attribute as a random variable with a certain hand-defined (conditional) probability distribution and relating them in a directed acyclic graph, we can use ancestral sampling~\cite{sam:Bishop07} to efficiently sample a diverse set of scene parameters $\saSim$.  Second, we render the scene defined by the parameters $\saSim$ into a semantic top-view $\bevmapSim$ with the same dimensions as $\bevmapReal$.  It is important to highlight that rendering is easy, compared to photo-realistic rendering of perspective RGB images~\cite{sam:Richter17a,sam:Richter16a}, because our model (i) works in the top-view where occlusion reasoning is not required and (ii) is defined in semantic space making illumination or photo-realism obsolete.  We generate a data set $\dsSim = \{\bevmapSim, \saSim\}_{i=1}^{\dsSimSize}$ of $\dsSimSize$ simulated semantic top-views $\bevmapSim$ and corresponding $\saSim$.
Fig.~\ref{fig:overview_scene_model} (right) gives a few examples of rendered top-views.


\subsection{Training and Inferring the Scene Model}
\label{sec:train_infer_scene_model}
We propose a deep learning framework that maps a semantic top-view $\bevmap$ into the scene model parameters $\sa$. Fig.~\ref{fig:framework_overview} provides a conceptual illustration.
To leverage both sources of supervision (real and simulated data) during training, we define this mapping as
\begin{equation}
    \sa = \nnFull(\bevmap) = (\nnAttr \circ \nnFeat)(\bevmap) \;,
\end{equation}
where $\circ$ defines a function composition and $\nnAttr$ and $\nnFeat$ are neural networks, with weights $\nnAttrWgts$ and $\nnFeatWgts$ respectively, that we want to train.
The architecture of $\nnFeat$ is a 6-layer convolutional neural network (CNN) that converts a semantic top-view $\bevmap \in \realspace^{\bevmapH \times \bevmapW \times \bevmapC}$ into a 1-dimensional feature vector $\bevmapFeat \in \realspace^{\bevmapFeatDim}$.
Then, the function $\nnAttr$ is defined as a multi-layer perceptron (MLP) predicting the scene attributes $\sa$ given $\bevmapFeat$.
Specifically, $\nnAttr$ is implemented as a multi-task network with three separate predictions $\predSaBin$, $\predSaMc$ and $\predSaReg$ for each of the parameter groups $\saBin$, $\saMc$ and $\saReg$ of the scene model.

Our objective is that $\nnAttr \circ \nnFeat$ works well on real data, while we want to leverage the rich and large set of annotations from simulated data during training.  The intuition behind our design is to have a unified $\nnFeat$ that maps semantic top-views $\bevmap$ of different domains into a common feature representation, usable by a domain-agnostic classifier $\nnAttr$.  To realize this intuition, we define supervised loss functions on both real and simulated data and leverage domain adaptation techniques to minimize the domain gap between the output of $\nnFeat$ given top-views from different domains.


\paragraph{Loss functions on scene attribute annotation:}
Given data sets $\dsReal$ and $\dsSim$ of real and simulated data, we define
\begin{equation}
    \lossSup = \lossWgtReal \cdot \lossSupReal + \lossWgtSim \cdot \lossSupSim
\end{equation}
as supervised loss.
The scalars $\lossWgtReal$ and $\lossWgtSim$ weigh the importance between real and simulated data and
\begin{equation}
\begin{split}
    \lossSupRealSim = \sum_{i=1}^{\dsRealSimSize} & \lossBCE(\saBinRealSimIdx, \predSaBinRealSimIdx) \\
                     & + \lossCE(\saMcRealSimIdx, \predSaMcRealSimIdx) \\
                     & + \lossEllOne(\saRegRealSimIdx, \predSaRegRealSimIdx) \;,
\end{split}
\end{equation}
where (B)CE is the (binary) cross-entropy loss and $\{\sa,\predSa\}_{\cdot,i}$ denotes the $i$-th sample in the data set.  For regression, we discretize continuous variables into $\numSaRegBins$ bins by convolving a dirac delta function centered at $\saReg$ with a Gaussian of fixed variance, which enables easier multi-modal predictions and is useful for the graphical model defined in Sec.~\ref{sec:graphical_model}.  We ignore scene attributes without manual annotation for $\lossSupReal$.

\begin{figure}
    \centering
    \includegraphics[width=1.0\columnwidth]{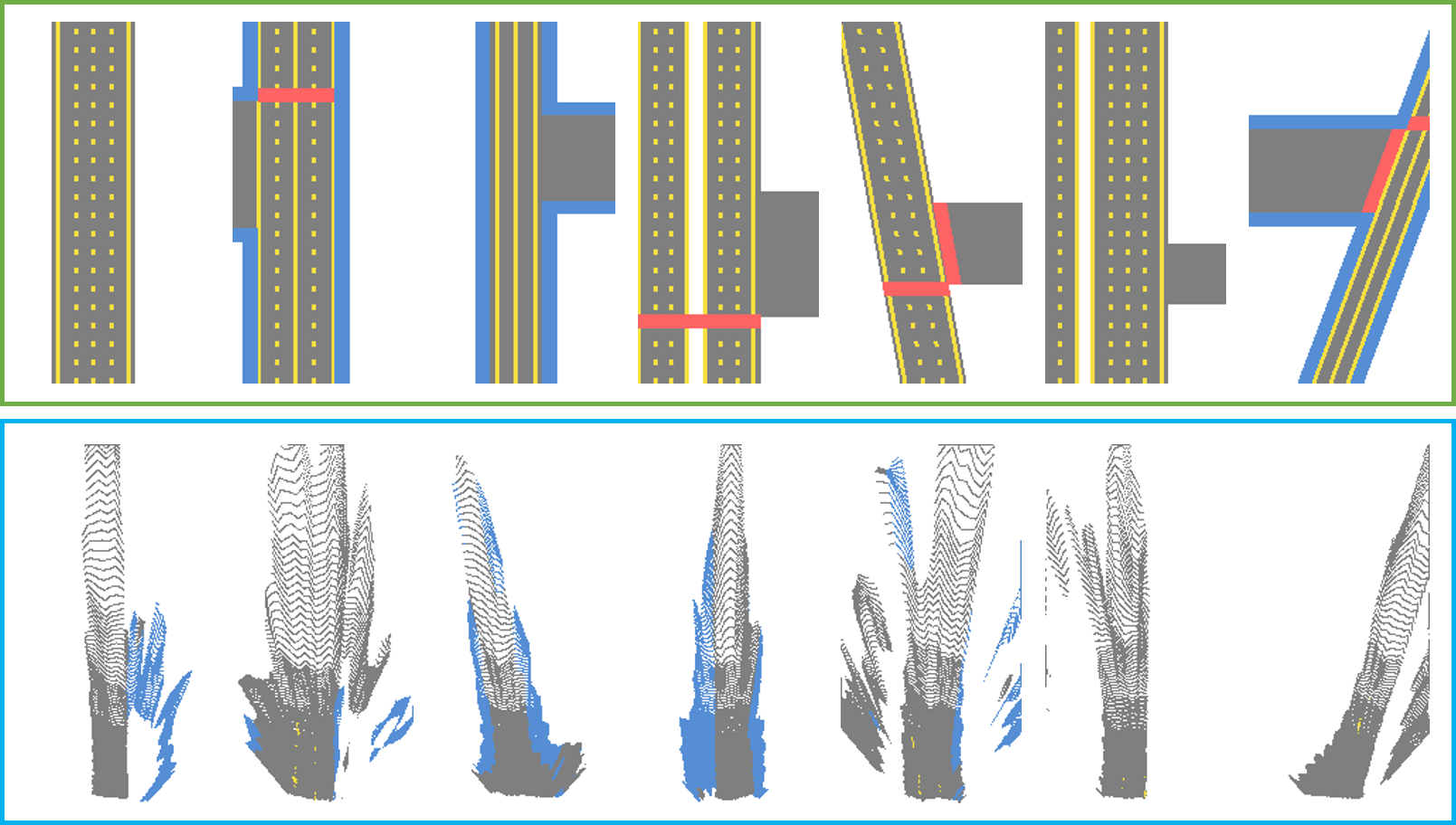}
    \vspace{-0.65cm}
    \caption{
    Unpaired examples of simulated semantic top-views (top) and real ones from \cite{sam:Schulter18a} (bottom). 
    }
    \label{fig:top_view_examples_real_and_sim}
\end{figure}

\paragraph{Bridging the domain gap:}
Since our goal is to leverage simulated data during the training process, our network design needs to account for the inherent domain gap.  We thus define separate feature extraction networks $\nnFeatReal$ and $\nnFeatSim$ with shared weights $\nnFeatWgts$ that take as input semantic top-views from either domain, \ie, $\bevmapReal$ or $\bevmapSim$, and compute respective features $\bevmapFeatReal$ and $\bevmapFeatSim$.  We then explicitly encourage a domain-agnostic feature representation by employing an adversarial loss function $\lossAdv$~\cite{sam:Ganin16a}.  We use an MLP $\nnDiscr(\bevmapFeat)$ with parameters $\nnDiscrWgts$ as discriminator, that takes the feature representations from either domain, \ie, $\bevmapFeatReal$ or $\bevmapFeatSim$, as input and makes a binary prediction into "real" or "fake".  As in standard generative adversarial networks, $\nnDiscr$ has the goal to discriminate between the two domains, while the rest of the model aims to confuse the discriminator by providing inputs $\bevmapFeatRealSim$ indistinguishable in the underlying distribution, \ie, a domain-agnostic representation of the semantic top-view maps $\bevmapRealSim$.
Fig.~\ref{fig:top_view_examples_real_and_sim} shows {\em unpaired} examples of simulated and real top-views to illustrate the domain gap.  Note that we chose domain adaptation on the feature-level over our initial attempt on the pixel-level with a modified version of~\cite{ziyan:zhu2017unpaired} due to a simpler design and higher accuracy.  However, we refer to the supplementary for a discussion of our pixel-level approach, which provides insights into the role of domain adaptation and further visualizations on overcoming the domain gap.

\paragraph{Optimization:}
We use ADAM~\cite{sam:Kingma15a} to estimate the parameters of our neural network model by solving:
\begin{equation}
  \max_{\nnDiscrWgts} \min_{\nnFeatWgts,\nnAttrWgts} \lossSup + \lossWgtAdv \lossAdv .
\end{equation}
Figure \ref{fig:framework_overview} provides an overview of our framework.

\subsection{CRF for Coherent Scene Understanding}
\label{sec:graphical_model}
We now introduce our graphical model for predicting consistent layouts of road scenes.  We first present our CRF for single frames and then extend it to the temporal domain.

\paragraph{Single image CRF:} Let us first denote the elements of scene attributes and corresponding predictions as $\sa[\cdot]$ and $\predSa[\cdot]$, where we use indices $i \in \{1, ..., \saBinNum\}$, $p \in \{1, ..., \saMcNum\}$ and $m \in \{1, ..., \saRegNum\}$ for binary, multi-class and continuous variables, respectively.
We then formulate scene understanding as the energy minimization problem
\begin{equation}
\begin{split}
    E(\sa|\bevmap) =& E_b(\saBin) + E_m(\saMc) + E_c(\saReg) \\
                    & + E_s(\saBin,\saMc)+ E_q(\saBin,\saReg) \\
                    & + E_h(\saBin,\saMc,\saReg) \;,
\end{split}
\label{eq:total_energy}
\end{equation}
where $E_{*}$ denotes energy potentials for the associated scene attribute variables ($\saBin$, $\saMc$ and $\saReg$).  We will describe the details for each of those potentials in the following.

For binary variables $\saBin$, our potential function $E_b$ consists of two terms,
\begin{equation}
  E_b(\saBin) = \sum_i \phi_b(\saBin[i]) + \sum_{i \neq j} \psi_b(\saBin[i],\saBin[j]) \;.
\end{equation}
The unary term $\phi_b(\cdot)$ specifies the cost of assigning a label to $\saBin^i$ and is defined as $-\log P_b(\saBin[i])$, where $P_b(\saBin[i]) = \predSaBin[i]$ is the probabilistic output of our neural network $\nnAttr$.
The pairwise term $\psi_b(\cdot,\cdot)$ defines the cost of assigning $\saBin[i]$ and $\saBin[j]$ to $i$-th and $j$-th variable as $\psi_b(\saBin[i],\saBin[j])=-\log M_b(\saBin[i],\saBin[j])$, where $M_b$ is the co-occurrence matrix and $M_b(\saBin[i],\saBin[j])$ is the corresponding probability.
For multi-class variables, our potential is defined as $E_m(\saMc)=\sum_p \phi_m(\saMc[p])$, where $\phi_m(\cdot) = -\log P_m(\cdot)$ and $P_m(\saMc[p]) = \predSaMc[p]$.
Similarly, we define the potential for continuous variables as $E_c(\saReg) = \sum_m \phi_c(\saReg[m])$ with $\phi_c(\saReg[m])$ being the negative log-likelihood of $\predSaReg[m]$.

For a coherent prediction, we further introduce the potentials $E_s$, $E_q$ and $E_h$ to model correlations among scene attributes.
$E_s$ and $E_q$ enforce hard constraints between certain binary variables and multi-class or continuous variables, respectively. They convey the idea that, for instance, the number of lanes of a side-road is consistent with the actual existence of that side-road.
We denote the set of pre-defined pairs between $\saBin$ and $\saMc$ as $\mathcal{S}=\{(i,p)\}$ and between $\saBin$ and $\saReg$ as $\mathcal{Q}=\{(i,m)\}$. 
Potential $E_s$ is then defined as
\begin{equation}
    E_s(\saBin,\saMc) = \sum_{(i,p) \in \mathcal{S}} \infty \times \mathbbm{1} \textlbrackdbl \saBin[i] \neq \saMc[p] \textrbrackdbl \;,
\end{equation}
where $\mathbbm{1}\textlbrackdbl * \textrbrackdbl$ is the indicator function.  Potential $E_q$ is defined likewise but using the set $\mathcal{Q}$ and variables $\saReg$.
In both cases, we give a high penalty to scenarios where two types of predictions are inconsistent.
   
Finally, the potential $E_h$ of our energy defined in Eq.~\eqref{eq:total_energy} models higher-order relations between $\saBin$, $\saMc$ and $\saReg$.
The potential takes the form
\begin{equation}
    E_h(\saBin,\saMc,\saReg) = \sum_{c\in \mathcal{C}} \infty \times f_c(\saBin[i],\saMc[p],\saReg[m]) \;,
\end{equation}
where $c=(i,p,m)$ and $f_c(\cdot,\cdot,\cdot)$ is a table where conflicting predictions are set to 1.
The supplementary gives a complete definition of $\mathcal{C}$ which contains the relations between scene attributes and the constraints we enforce on them.

\paragraph{Temporal CRF:}
Given videos as input, we propose to extend our CRF to encourage temporally consistent and meaningful outputs.
We extend the energy function from Eq.~\eqref{eq:total_energy} by two terms that enforce temporal consistency of binary and multi-class variables and smoothness for continuous variables.  Due to space limitations, we refer to the supplementary for details of our formulation.

\paragraph{Learning and inference on CRF:}
We use QPBO~\cite{rother2007optimizing} for inference in both CRF models.
Since ground truth is not available for all frames, we do not introduce per-potential weights.
However, our CRF is amenable to piece-wise~\cite{sutton2005piecewise} or joint learning~\cite{domke2013learning,zheng2015conditional} if ground-truth is provided.


%% file: sections/experiments.tex
\section{Experiments}
\label{sec:exps}
To evaluate the quality of our scene understanding approach we conduct several experiments and analyze the importance of different aspects of our model.
Since we do have manually-annotated ground truth, we can quantify our results and compare with several baselines that demonstrate the impact of two key contributions: the use of top-view maps and simulated data for training.
We also put an emphasis on qualitative results in this work for two reasons:
First, not all attributes of our model are actually contained in the manually-annotated ground truth and can thus not be quantified but only qualitatively verified.
Second, there is obviously no prior art showing results on this novel set of ground truth data, which makes the analysis of qualitative results even more important.

\paragraph{Datasets:}
Since our focus is on driving scenes and our approach requires semantic segmentation and depth annotation, we choose to work with the KITTI~\cite{sam:Geiger13a} and the newly released NuScenes~\cite{sam:NuScenes18a}\footnote{At the time of conducting experiments, we only had access to the pre-release of the data set.} data sets.
Although both data sets provide laser-scanned data for depth ground truth, note that depth supervision can also come from stereo images~\cite{sam:Godard17a}.
Also, since NuScenes~\cite{sam:NuScenes18a} does not provide semantic segmentation, we reuse the segmentation model from KITTI.
For both data sets, we manually annotate a subset of the images with our scene attributes.
Annotators see the RGB image as well as the depth ground truth and provide labels for 22 attributes of our model.
We refer to the supplementary for details on the annotation process.
In total, we acquired around 17000 annotations for KITTI~\cite{sam:Geiger13a} and 3000 for NuScenes~\cite{sam:NuScenes18a}.

\paragraph{Evaluation metrics:}
Since the output space of our prediction is complex and consists of a mixture of discrete and continuous variables, which require different handling, we use multiple different metrics for evaluation.

For binary variables (like the existence of side roads) and for multi-class variables (like the number of lanes), we measure accuracy as $\textrm{Accu.-Bi} = \frac{1}{14} \sum_{k=1}^{14} [p_k = \saBin{}_k]$ and $\textrm{Accu.-Mc} = \frac{1}{2} \sum_{k=1}^{2} [p_k = \saMc{}_k]$.
For regression variables we use the mean squard error (MSE).  

Besides these standard metrics, we also propose another metric that combines all predicted variables and outputs into a single number.
We take the predicted parameters and render the scene accordingly.
For the corresponding image, we take the ground truth parameters (augmented with predicted values for variables without ground truth annotation) and render the scene, which assigns each pixel a semantic category.
For evaluation, we can now use Intersection-over-Union (IoU), a standard measure in semantic segmentation.
While being a very challenging metric in this setup, it implicitly weighs the attributes by their impact on the area of the top-view.
For instance, predicting the number of lanes incorrectly by one has a bigger impact than getting the distance to a sideroad wrong by one meter.

\begin{figure*}\centering
\includegraphics[width=1.0\textwidth]{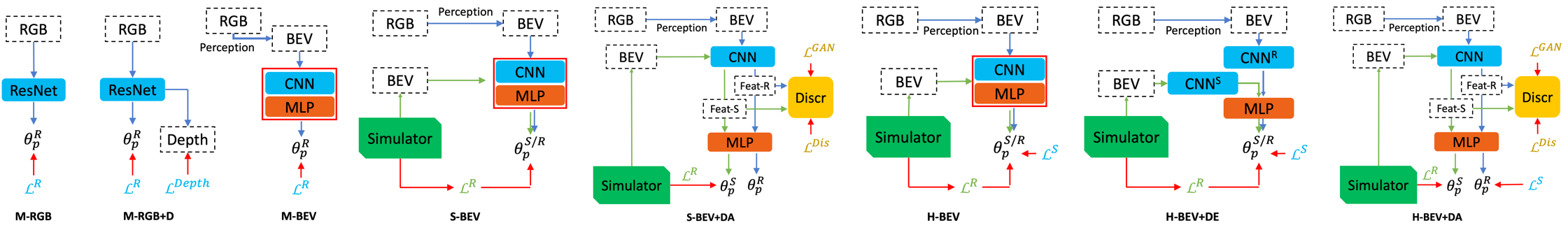}
\vspace{-0.65cm}
\caption{Illustrations of all the models we compare in the quantitative evaluation in Tab.~\ref{tbl:main_exp_layout}.}
\label{fig:archs_overview}
\end{figure*}

\begin{table*}\centering\small
  \input{tables/main_exp_layout}
  \vspace{-0.2cm}
  \caption{Main results on road scene layout estimation on both data sets KITTI~\cite{sam:Geiger13a} and NuScenes~\cite{sam:NuScenes18a}.}
  \label{tbl:main_exp_layout}
\end{table*}

\subsection{Single Image Evaluation}
Our main experiments are conducted with a single image as input.  In the next section, we separately evaluate the impact of temporal modeling as described in Sec.~\ref{sec:graphical_model}.

\paragraph{Baselines:}
Since we propose a scene model of roads with new attributes and corresponding ground truth annotation, there exist no previously reported numbers.
We thus choose appropriate baselines that are either variations of our model or relevant prior works extended to our scene model:
\begin{itemize}
\item \textbf{Manual-GT-RGB (M-RGB):} A classification CNN (ResNet-101~\cite{sam:He16a}) trained on the manually-annotated ground truth.  Seff and Xiao~\cite{sam:Seff16a} have the same setup except that we use a network with more parameters and train for all attributes simultaneously in a multi-task setup.
\item \textbf{Manual-GT-RGB+Depth (M-RGB+D):} Same as M-RGB but with the additional task of monocular depth prediction (as in our perception model).  The intuition is that this additional supervision aids predicting certain scene attributes, \eg, distances to side roads, and renders a more fair comparison point to our model.
\item \textbf{Manual-GT-BEV (M-BEV):} Instead of using the perspective RGB image as input, this baseline uses the output of~\cite{sam:Schulter18a}, which is a semantic map in the top-view, also referred to as bird's eye view (BEV).
We train the function $\nnFull$ with the manually annotated ground truth.  Thus, M-BEV can be seen as an extension of~\cite{sam:Schulter18a} to our scene model.
\item \textbf{Simulation-BEV (S-BEV):} This baseline uses the same architecture as M-BEV but is trained only in simulation.
\item \textbf{Simulation-BEV+DomainAdapt (S-BEV+DA):} Same as S-BEV, but with additional domain adaptation loss as proposed in our model.
\end{itemize}
We denote our approach proposed in Sec.~\ref{sec:method}, according to the nomenclature above, as \textbf{Hybrid-BEV+DomainAdapt (H-BEV+DA)} and further explore two variants of it.
First, \textbf{H-BEV} does not employ the discriminator $\nnDiscr$ but still trains from both domains.
Second, \textbf{H-BEV+DE} also avoids the discriminator but uses a separate set of weights $\nnFeatWgtsReal$ and $\nnFeatWgtsSim$ for the feature extraction network $\nnFeat$.
The intuition is that the supervised losses from both domains and the separate domain-specific encoding (thus, "+DE") already provide enough capacity and information to the model to find a domain-agnostic representation of the data.
Please refer to Fig.~\ref{fig:archs_overview} for an overview of the different models we compare.
For the best models among each group (M-, S- and H-), we report numbers with the graphical model (+GM).

\paragraph{Quantitative results:}
Tab.~\ref{tbl:main_exp_layout} summarizes our main results for both data sets and we can draw several conclusions.
First, when comparing the groups of methods by supervision type, \ie, manual (M), simulation (S) and hybrid (H), we can clearly observe the benefit of hybrid methods leveraging both domains.  Second, within the group of manual annotation, we can see that adding depth supervision to the approach of~\cite{sam:Seff16a} significantly improves results, particularly for continuous variables.  Predicting the scene attributes directly from the top-view representation of~\cite{sam:Schulter18a} is slightly better than M-RGB+D on KITTI and worse on NuScenes, but has the crucial advantage that augmentation with simulated data in the top-view becomes possible, as illustrated with all hybrid variants.  Third, within the group of simulated data, using domain adaptation techniques (S-BEV+DA) has a significant benefit.  We want to highlight the competitive overall results of S-BEV+DA, which is an unsupervised domain adaptation approach requiring no manual annotation.  Forth, also for hybrid methods, explicitly addressing the domain gap (H-BEV+DE and H-BEV+DA) enables higher accuracy.  Finally, all models improve with our graphical model put on top.

\begin{figure*}
	\centering
	\begin{tabular}{ccc}
		\includegraphics[width=0.31\textwidth]{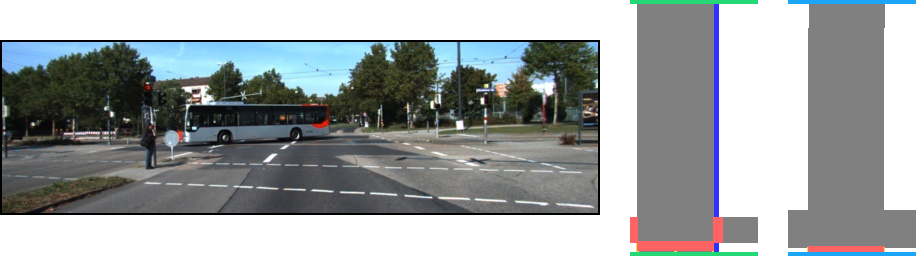} 
		& \includegraphics[width=0.31\textwidth]{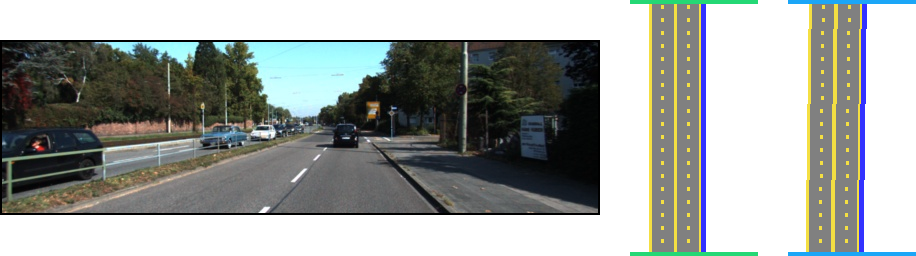}
		&\includegraphics[width=0.31\textwidth]{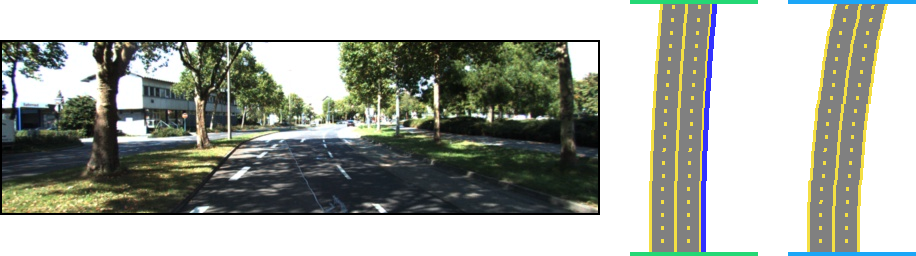} \\
		\includegraphics[width=0.31\textwidth]{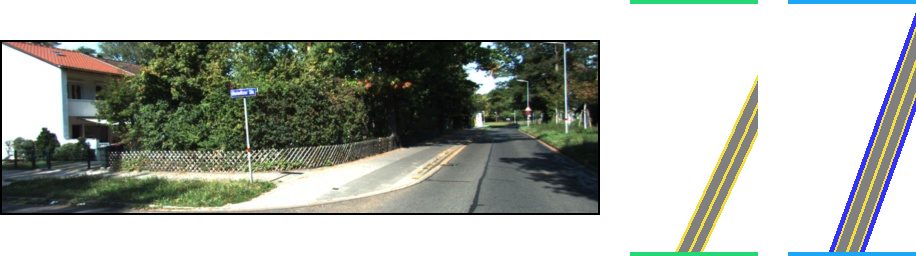}
		&\includegraphics[width=0.31\textwidth]{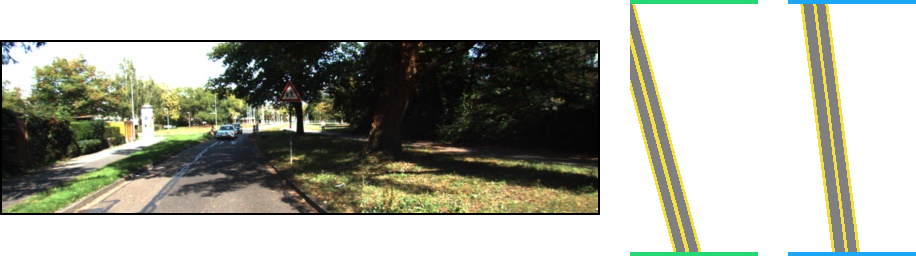} 
		& \includegraphics[width=0.31\textwidth]{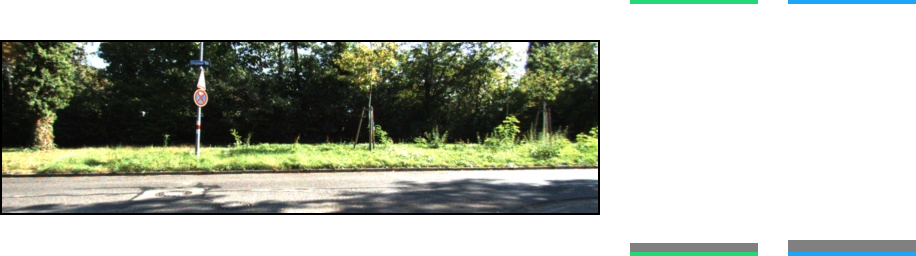} \\
		\includegraphics[width=0.31\textwidth]{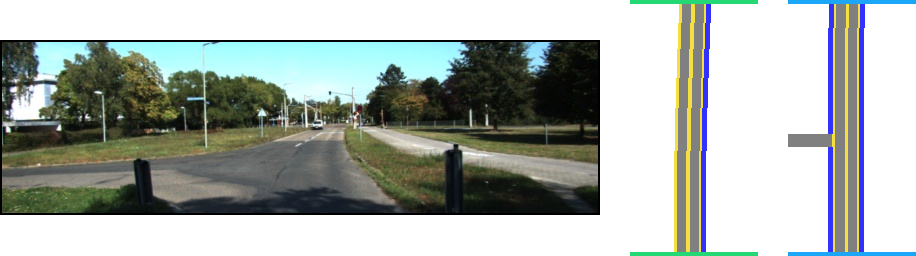}
		& \includegraphics[width=0.31\textwidth]{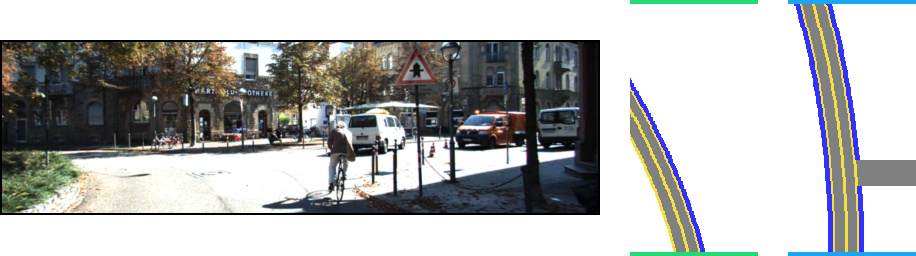}
		& \includegraphics[width=0.31\textwidth]{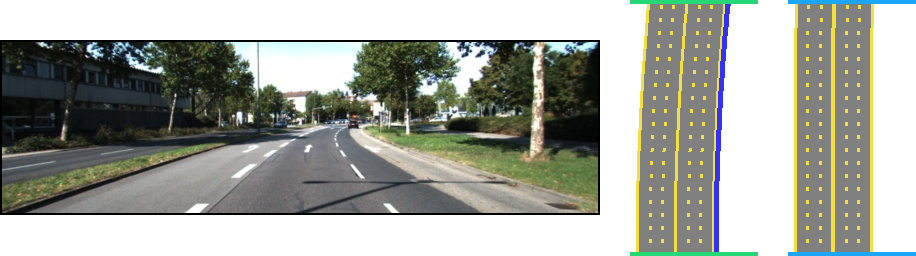} \\
	\end{tabular}
	\vspace{-0.35cm}
	\caption{Qualitative results of H-BEV+DA+GM on individual frames from KITTI.  Each example shows perspective RGB, ground truth and predicted semantic top-view, respectively.  Our representation is rich enough to cover various road layouts and handles complex scenarios, \eg, rotation, existence of crosswalks, sidewalks, side-roads and curved roads.}
	\label{fig:GM_output_more}
\end{figure*}

\begin{figure*}
	\centering
	\begin{tabular}{ccc}
	\includegraphics[width=0.31\textwidth]{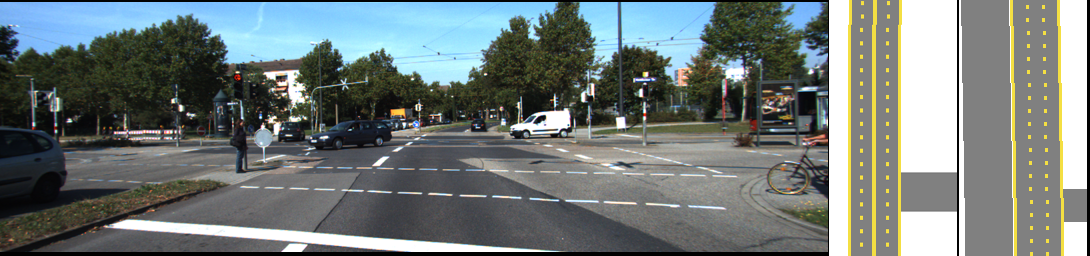} 
	& \includegraphics[width=0.31\textwidth]{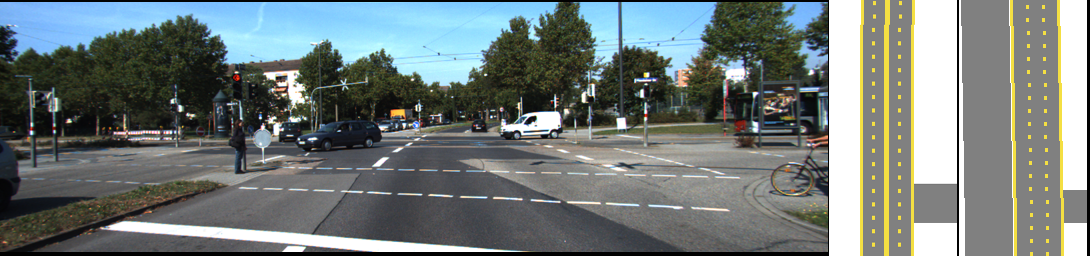} 
	& \includegraphics[width=0.31\textwidth]{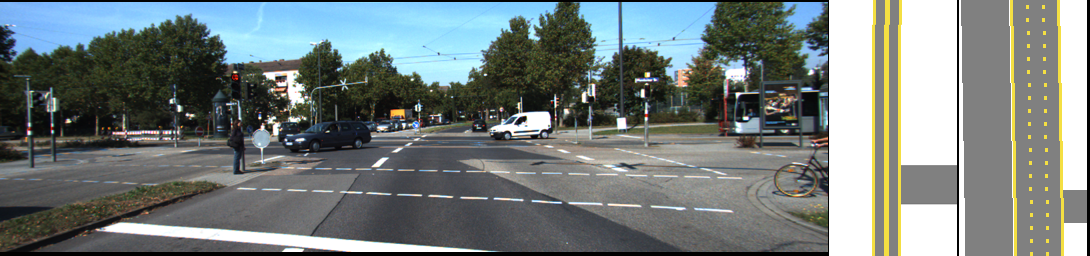} \\
	\includegraphics[width=0.31\textwidth]{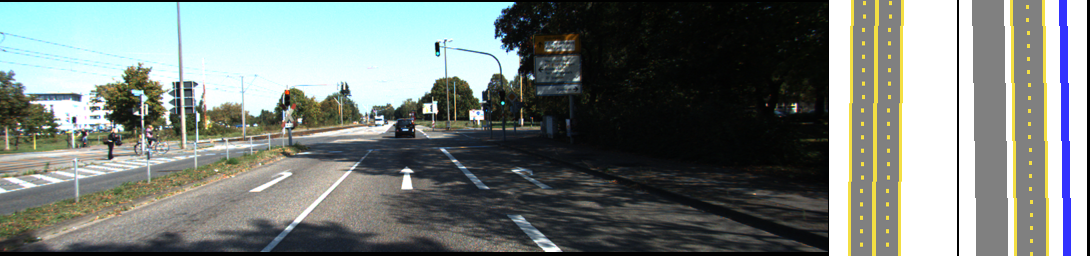} 
	& \includegraphics[width=0.31\textwidth]{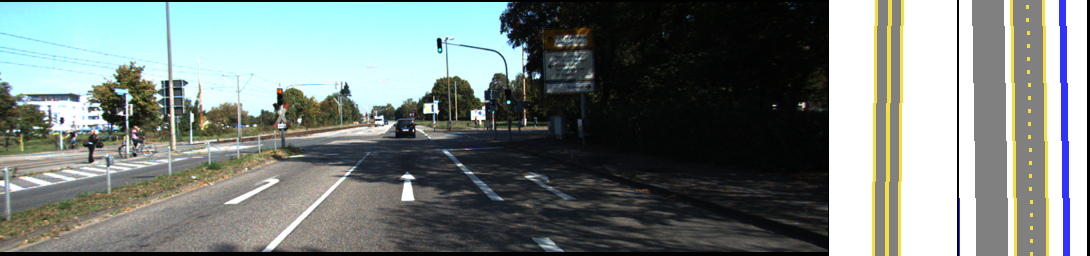} 
	&\includegraphics[width=0.31\textwidth]{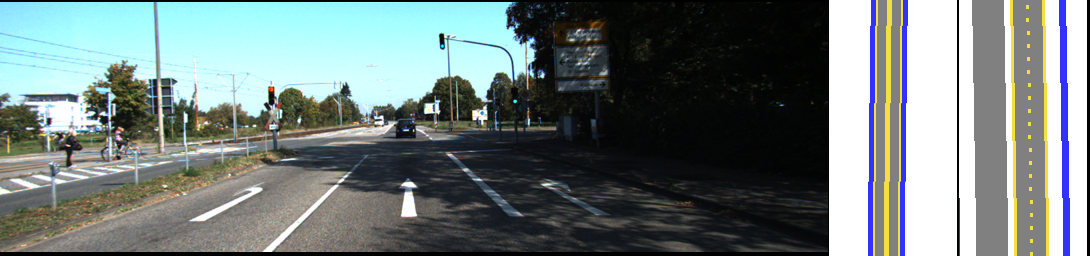} \\
	\end{tabular}
	\vspace{-0.35cm}
	\caption{Qualitative results comparing H-BEV+DA and H-BEV+DA+GM in consecutive frames of two example sequences of the KITTI validation set. In each column, we have visualized the perspective RGB image, prediction from H-BEV+DA and that of H-BEV+DA+GM from left to right. Each row shows a sequence of three frames. We can observe more consistent predictions, \eg, width of side-road and delimiter width, with the help of the temporal CRF.}
	\label{fig:GM_output}
\end{figure*}

\paragraph{Qualitative results:}
We show several qualitative results in Fig.~\ref{fig:GM_output_more} and Fig.~\ref{fig:GM_output} and again highlight their importance to demonstrate the practicality of our approach qualitatively.  We can see from the examples that our model successfully describes a diverse set of road scenes.

\begin{table}\centering\small
  \input{tables/consistency_check}
  \vspace{-0.2cm}
  \caption{Main results on consistency measurements.}
  \label{tbl:consistency_table}
\end{table}

\subsection{Evaluating consistency of our model}
We now analyze the impact of the graphical model on the consistency of our predictions, for which we define the following metrics:

\begin{itemize}
	\item {\em Semantic consistency:} we measure the conflicts in attribute predictions w.r.t. their semantic meanings.  Specifically, we count a conflict if predicted attributes are not feasible in our scene model. The average number of conflicts is reported as our semantic consistency measurement.
	\item {\em Temporal consistency:} for each attribute prediction among a video sequence, we measure the number of changes in the prediction. We report the average number of prediction changes as the temporal consistency. The lower the number is, the more stable prediction we would obtain. Note that consistency itself cannot replace the accuracy since a prediction can also be consistently wrong. 
\end{itemize}

As for the temporal consistency, we visualize qualitative results of consecutive frames in two validation sequences from KITTI in Fig.~\ref{fig:GM_output}.  The graphical model successfully enforces temporal smoothness, especially for number of lanes, delimiter width and the width of side-roads.

Finally, we show in Tab.~\ref{tbl:consistency_table} quantitative results for the temporal consistency metrics defined above on both KITTI and NuScenes data sets.  We compare representative models from each group of different forms of supervision (M-, S- and H-) with the output of the graphical model applied on H-BEV+DA.  We can clearly observe a significant improvement in consistency for both data sets.  Together with the superior results in Tab.~\ref{tbl:main_exp_layout}, this clearly demonstrates the benefits of the proposed graphical model for our application.
 


%% file: tables/main_exp_layout.tex
\begin{tabular}{l|cccc|ccccc}
  \hline
              &  \multicolumn{4}{c|}{KITTI~\cite{sam:Geiger13a}} & \multicolumn{4}{c}{NuScenes~\cite{sam:NuScenes18a}} \\
  \hline
  Method      &  Accu.-Bi. $\uparrow$ & Accu.-Mc. $\uparrow$ & MSE $\downarrow$ & IOU $\uparrow$   & Accu.-Bi. $\uparrow$ &  Accu.-Mc. $\uparrow$ & MSE $\downarrow$ & IOU $\uparrow$ \\
  \hline\hline
  M-RGB~\cite{sam:Seff16a}          & .811& .778& .230& .317& .846& .604& .080& .316\\
  M-RGB~\cite{sam:Seff16a}+D        & .799& .798& .146& .342& .899 & .634& .021& .335 \\
  M-BEV~\cite{sam:Schulter18a}      & .820& .777& .141& .345& .852& .601& .022&.269 \\
  M-BEV~\cite{sam:Schulter18a} +GM  & .831& .792& .136& .350& .852 & .601 & .036 & .338 \\
  \hline
  S-BEV                             & .694& .371& .249& .239 & .790& .366& .162&.155 \\
  S-BEV+DA                          & .818& .677& .222& .314& .753& .568& .103&.171 \\
  S-BEV+DA+GM                       & .847& .683& .230& .320& .723 & .568 & .081 & .160 \\
   \hline
  H-BEV                             & .816 & .756 & .152 & .342 & .783 & .569 & .039 & .345 \\
  H-BEV+DE                          & .830 & .776 & .158 & .381 & .854 & .626 & .042 & .423 \\
  H-BEV+DA                          & .845 & .792 & .108 & .398 & .856 & .545 & .028 & .346 \\
  H-BEV+DA+GM                       & .849 & .805 & .098 & .371 & .855 & .626 & .033 & .450 \\
  \hline
\end{tabular}


%% file: tables/consistency_check.tex
\begin{tabular}{l|cc|cc}
  \hline
              &  \multicolumn{2}{c|}{KITTI~\cite{sam:Geiger13a}} & \multicolumn{2}{c}{NuScenes~\cite{sam:NuScenes18a}} \\
  \hline 
  Method      &  seman.$\downarrow$& temp.$\downarrow$ & seman.$\downarrow$ & temp.$\downarrow$ \\ 
  \hline
  S-BEV+DA & 2.82& 5.32& 1.08 & 2.09 \\
  M-BEV~\cite{sam:Schulter18a} & 2.65& 3.99& 1.09 & 1.27\\
  H-BEV+DA & 5.59& 6.01& 1.08 & 1.05 \\
  \hline
  \hspace{28pt}+GM& 1.77& 1.93&  0.11 & 0.42\\                  
  \hline
\end{tabular}

%% file: sections/conclusion.tex
\section{Conclusion}
\label{sec:conclusion}
In this work, we present a scene understanding framework for complex road scenarios. Our key contributions are:  (1) A parameterized and interpretable model of the scene that is defined in the top-view and enables efficient sampling of diverse scenes.  The semantic top-view representation makes rendering easy (compared to photo-realistic RGB images in perspective view), which enables the generation of large-scale simulated data.  (2) A neural network design and corresponding training scheme to leverage both simulated as well as manually-annotated real data.  (3) A graphical model that ensures coherent predictions for a single frame input and temporally smooth outputs for a video input.
Our proposed hybrid model (using both sources of data) outperforms its counterparts that use only one source of supervision in an empirical evaluation. This confirms the benefits of the top-view representation, enabling simple generation of large-scale simulated data and consequently our hybrid training.

\paragraph{Acknowledgements:} We want to thank Kihyuk Sohn for valuable discussions on domain adaptation and all anonymous reviewers for their comments.
